\documentclass{article}
\usepackage[utf8]{inputenc}
\usepackage{graphicx}
\usepackage{natbib}
\usepackage{xurl}

\usepackage{hyperref}
\usepackage{authblk}
\usepackage{csquotes}
\usepackage{authblk}

\usepackage{xcolor}
\usepackage{hyperref}
\hypersetup{
    colorlinks=true,
    linkcolor=blue,      
    citecolor=blue,     
    filecolor=blue,   
    urlcolor=blue       
}
\hypersetup{
  colorlinks=true,
  linkcolor=myblue,    
}

\title{Navigating News Narratives: A Media Bias Analysis Dataset}
\author{Shaina Raza}
\affil{Vector Institute for Artificial Intelligence}
\affil{Email: \texttt{shaina.raza@vectorinstitute.ai}}
\date{}

\begin{document}

\maketitle

\begin{abstract}
The proliferation of biased news narratives across various media platforms has become a prominent challenge, influencing public opinion on critical topics like politics, health, and climate change. This paper introduces ``Navigating News Narratives: A Media Bias Analysis Dataset,'' a comprehensive dataset to address the urgent need for tools to detect and analyze media bias. This dataset encompasses a broad spectrum of biases, making it a unique and valuable asset in the field of media studies and artificial intelligence.

The dataset is available at \href{https://huggingface.co/datasets/newsmediabias/news-bias-full-data}{Dataset page}.

\end{abstract}

\section{Introduction}
In an era where information is ubiquitous, the news media's role expands beyond mere reporting; it actively constructs and frames public discourse \citep{raza_news_2022}. The power of the media to influence perception and decision-making cannot be understated, especially in a world where news is consumed in real-time from a number of sources \citep{raza_automatic_2021}. Traditional notions of media as impartial observers are giving way to a more critical understanding of its participatory role in shaping socio-political narratives.

The dataset introduced in this research represents a crucial step in addressing the multifaceted nature of news media bias. This data includes a wide array of dimensions such as \textit{race, gender, age, occupation, and climate change}. The dataset provides a holistic tool for showing the complex interplay of factors that characterize contemporary news media. This approach acknowledges that biases in media are not monolithic but are instead a confluence of various underlying factors, each contributing to the overarching narrative in its unique way \citep{lei-etal-2022-sentence}.

This expansive scope is particularly important given the current global landscape, where issues of racial and gender equality, climate change, and political polarization are at the forefront of public consciousness. Media narratives around these issues not only inform public opinion but also shape policy and societal norms \citep{raza_fairness_2023}. Thus, understanding and identifying biases within these narratives is not merely an academic exercise; it is a societal imperative.

The methodology employed in creating this dataset leverages active learning \citep{xiao_freeal_2023}. Active learning involves training models on a progressively expanding dataset where the model itself identifies the most informative data points. By combining manual labeling with semi-supervised learning and iterative human verification, the dataset achieves a high degree of accuracy and reliability \citep{raza_constructing_2023}. 

We present this data to researchers, journalists, and policymakers  to understand and mitigate the influence of biased narratives in shaping public discourse. We welcome the community to contribute and make this dataset as effort in navigating the complex world of news narratives.

\section{Dataset Description}
\subsection{Data Composition}
The dataset comprises various categories of news, including climate crisis news, occupational news, and spiritual news, among others. The key attributes of the dataset include:

\begin{itemize}
    \item \textit{ID}: A unique numeric identifier for each news item.
    \item \textit{Text:} The main content of the news item.
    \item \textit{Dimension:} A categorical descriptor indicating the type of bias.
    \item \textit{Biased\_Words}: A list of words identified as biased within the text.
    \item \textit{Aspect: }The specific topic or aspect covered in the text.
    \item \textit{Label:} An indicator of the level of bias, categorized as Neutral, Slightly Biased, or Highly Biased.
\end{itemize}

\subsection{Annotation Scheme}
The dataset employs an active learning-based annotation scheme, encompassing manual labeling, semi-supervised learning, and iterative human verifications \citep{raza_constructing_2023}. Active learning in this dataset is implemented in a multi-stage process. Initially, a subset of the data is manually labeled, providing a baseline of categorized examples. These initial labels cover a range of biases across different dimensions, setting a foundation for the model to understand the complexities of biased narratives. Following this, the dataset is introduced to a semi-supervised learning model, which begins to classify unlabeled data. The unique advantage of this step is the model's ability to learn from a combination of labeled and unlabeled data, enhancing its understanding of biases in a more organic and expansive manner.

As the model processes and classifies the data, its predictions are then subjected to iterative human verification. This step is crucial in maintaining the integrity of the dataset. Human reviewers assess the model's classifications, correcting errors and refining the model's criteria for bias identification. This iterative process not only ensures the accuracy of the dataset but also allows the model to adapt and improve over time, learning from the nuances and subtleties identified by human experts.

The combination of manual labeling, semi-supervised learning, and human verification culminates in a dataset of exceptional accuracy and reliability. This methodological rigor not only ensures that the dataset is comprehensive in its coverage but also precise in its analysis. It allows for the detailed examination of media bias across various dimensions, making it an invaluable tool for researchers and practitioners in the field of media studies and artificial intelligence.

\subsection{Data Sources}
This dataset has been enriched by integrating data from various reputable sources. 
\begin{itemize}
    \item MBIC (media bias) by \citep{spinde2021neural}
    \item Hyperpartisan news detection data by \citep{kiesel2019semeval}
    \item Toxic comment classification challenge data from Kaggle\citep{jigsaw-toxic-comment-classification-challenge} .
    \item Jigsaw Unintended Bias in Toxicity Classification  \citep{jigsaw-multilingual-toxic-comment-classification}
    \item Age Bias and Sentiment Analysis data by \citep{diaz2018addressing}
    \item News bias data related to Ukraine by \citep{Farber2020}
    \item Social bias frames by \citep{sap2019social}
\end{itemize}
 
We also \textbf{curated the data }by leveraging Google RSS feeds, utilizing a strategic approach to gather relevant news content. This process involved the use of specific keywords to filter and extract news articles from the time period of 2022-2023. The keywords were carefully selected to encompass a broad range of topics and biases, ensuring that the dataset reflects the diverse nature of media narratives.  These keywords are:

\begin{enumerate}
    \item \textit{Political Keywords}: Terms related to significant political events, parties, or figures relevant in 2022-2023. Examples might include "elections," "democracy," "authoritarianism," or the names of prominent political leaders or parties active during this period.
    \item \textit{Social Issues:} Keywords related to major social movements or issues. For instance, "Black Lives Matter," "gender equality," "LGBTQ rights," "immigration," or "gun control."
    \item \textit{Public Health: }Given the ongoing impact of the COVID-19 pandemic, terms like "COVID-19," "vaccine," "public health," or "pandemic response" could be relevant.
    \item \textit{Economic Terms}: Keywords reflecting economic concerns such as "inflation," "unemployment," "recession," "trade war," or specific policy terms like "stimulus package."
    \item \textit{Environmental Issues:} With climate change being a critical global issue, terms like "climate change," "global warming," "sustainable energy," or specific events like "COP27" (if it were held in 2023).
    \item \textit{International Relations:} Keywords related to major international events or tensions, such as "Ukraine," "Middle East," "NATO," "United Nations," or "trade agreements."
    \item \textit{Technology and Cybersecurity:} Terms like "artificial intelligence," "data privacy," "cybersecurity," or names of major tech companies.
    \item \textit{Cultural and Entertainment: }Keywords related to significant cultural events or figures, including terms like "Oscars," "Grammys," "sports championships," or the names of influential cultural figures.
    \item \textit{Misinformation/Disinformation:} Given the focus on media bias, including terms like "fake news," "propaganda," "media bias," or "fact-checking."
    \item \textit{Human Rights: }Keywords associated with human rights issues, such as "refugee crisis," "human trafficking," "freedom of speech," or "political asylum."
\end{enumerate}
These keywords should be dynamically adjusted to capture the evolving nature of news narratives and ensure the dataset remains representative of the current media landscape. Additionally, the selection of keywords must be carefully balanced to avoid introducing additional biases into the dataset.

 The dataset is available at:
\begin{itemize}
    \item \url{https://huggingface.co/datasets/newsmediabias/news-bias-full-data}
    \item \url{https://zenodo.org/records/10231028}
\end{itemize}
\subsection{Research Potential}

\subsubsection{AI and Machine Learning}

The dataset offers a rich resource for developing and testing AI algorithms aimed at detecting and analyzing media bias. Its diverse range of bias dimensions and detailed annotation scheme make it suitable for training sophisticated machine learning models.

\subsubsection{Media Studies}

Researchers in media  and communication studies can use this dataset to quantitatively analyze trends in media bias, providing insights into how news narratives may influence public opinion and societal norms.

\subsubsection{Ethical Implications}

This work also contributes to the broader discourse on ethical AI and responsible journalism, underscoring the need for transparency and fairness in news reporting and AI-driven media analysis.

\subsection{Conclusion}

``Navigating News Narratives: A Media Bias Analysis Dataset" is a significant contribution to the fields of media studies, artificial intelligence, and ethics. It provides a nuanced and comprehensive tool for analyzing media bias, facilitating a deeper understanding of how news narratives shape public discourse. This open-access dataset, licensed under CC BY-NC 4.\citep{ccbync40}, is expected to be a valuable resource for researchers, AI practitioners, and journalists worldwide.
\textbf{Please cite this dataset in your work if you use it.}
\section*{Acknowledgments}
We would like to acknowledge the support and facilitation provided by the Vector Institute for Artificial Intelligence.

\bibliographystyle{plainnat}
\bibliography{references} 

\end{document}